\documentclass{article}

\usepackage[preprint]{neurips_2024}

\usepackage[utf8]{inputenc} 
\usepackage[T1]{fontenc}    
\usepackage{hyperref}       
\usepackage{url}            
\usepackage{booktabs}       
\usepackage{amsfonts}       
\usepackage{nicefrac}       
\usepackage{microtype}      
\usepackage{xcolor}         
\usepackage{natbib}
\usepackage{graphicx}
\usepackage{amsmath}
\usepackage{amsthm}
\usepackage[switch]{lineno}
\usepackage{balance}
\usepackage{amssymb}
\usepackage{subcaption}

\title{Unraveling Token Prediction Refinement and Identifying Essential Layers in Language Models}

\author{%
  Jaturong Kongmanee\\
  Independent Researcher\\
  Toronto, Canada \\
  \texttt{dillsunnyb11@gmail.com} \\
}

\begin{document}

\maketitle

\begin{abstract}
This research aims to unravel how large language models (LLMs) iteratively refine token predictions through internal processing. We utilized a logit lens technique to analyze the model's token predictions derived from intermediate representations. Specifically, we focused on (1) how LLMs access and utilize information from input contexts, and (2) how positioning of relevant information affects the model's token prediction refinement process. On a multi-document question answering task with varying input context lengths, we found that the depth of prediction refinement (defined as the number of intermediate layers an LLM uses to transition from an initial correct token prediction to its final, stable correct output), as a function of the position of relevant information, exhibits an approximately inverted U-shaped curve. We also found that the gap between these two layers, on average, diminishes when relevant information is positioned at the beginning or end of the input context. This suggested that the model requires more refinements when processing longer contexts with relevant information situated in the middle. Furthermore, our findings indicate that not all layers are equally essential for determining final correct outputs. Our analysis provides insights into how token predictions are distributed across different conditions, and establishes important connections to existing hypotheses and previous findings in AI safety research and development.
\end{abstract}


\section{Introduction}
Recent advances in technologies have seen an increasing number of applications of capable AI systems in various domains. One example is that of large language models (LLMs) that have exhibited improved capabilities in content and code generation and language translation. As LLMs rapidly advance in sophistication and generality, understanding them is essential to ensure their alignment with human values and prevent catastrophic outcomes \citep{hendrycks_xrisk_2022,hendrycks_overview_2023,ji_ai_2024}. Research that aims to improve an understanding of LLM goes beyond simple performance metrics, to unravel the internal workings of LLMs. This work utilized an observational approach that analyzes the inner workings of neural networks: \textit{Logit Lens} (as discussed in detail in Section \ref{logit-lens}). We focused on how LLMs access and use information from input contexts, and how positioning of relevant information affects the model's token prediction refinement process, to gain more insights that potentially aid the development of methods that can ensure trustworthiness in LLMs.

Mechanistic interpretability (MI) is a growing approach that aims to precisely define computations within neural networks. It seeks to fully specify a neural network’s computations, aiming for a granular understanding of model behavior, akin to reverse-engineering the model’s processes into pseudo-code. This approach stands out for its ambitious goal of complete reverse engineering and its focus on AI safety. (for a list of introduction and comprehensive review, see \citep{olah_mechanistic_2022, nanda_comprehensive_2022, olah_zoom_2020, sharkey_current_2022, olah_building_2018, nanda_mechanistic_2023, nanda_extremely_2024}.)

MI uses a variety of tools, from observational analysis to causal interventions (for a more detailed discussion of taxonomies of explainable artificial intelligence and MI methods, see \citep{10-das2020taxonomy,11-speith2022review,12-bereska2024mechanistic,13-zytek2022need,14-zhang2021survey}). Causal methods span from purely observational, which analyze existing representations without manipulation, to interventional approaches that directly perturb model components with the hope to reveal causal relationships. In terms of learning phases, post-hoc techniques are applied after training, while intrinsic methods are designed to enhance interpretability during the training process itself. This is to gain a deeper understanding of the behavior of the model from the beginning.

Another approach, given the scope of analysis and application, is to categorize methods based on general tendencies of interpretability. Some can provide both local and global insight, or partial and comprehensive understanding. Probing techniques \citep{07-alain2016understanding, 05-ettinger2016probing, 06-hupkes2018visualisation}, for instance, range from local to global: simple linear probes reveal individual feature insights, while advanced structured probes \citep{18-burns2022discovering} expose broader patterns. Sparse autoencoders \citep{17-cunningham2023sparse}, while decomposing individual neuron activations (local), aim to disentangle features throughout the model (global). Path patching \citep{15-wang2022interpretability,16-goldowsky2023localizing} transforms local interventions into global understanding by tracing information flow across layers, showing how small perturbations can illuminate the entire model's behavior.

In this work, \textit{logit lens} technique--one among the other observational methods--was employed (1) to examine how network layers iteratively refine token predictions, and (2) to observe its relationships to \textit{input context lengths} and \textit{the position of relevant information within input context}, and to another method under the same category: \textit{probing}. In the remainder of this paper, details about logit lens technique is provided in Section \ref{logit-lens}. Section \ref{input-context-logit-lens} presents the relationships between logit lens and input context lengths and the position of relevant information within the input context. Relation of logit lens to probing is discussed in Section \ref{probing-logit-lens}. Section \ref{ai-safety} gives an overview of how understanding inner working mechanisms of LLM is relevant to AI safety research and development.

\section{Logit Lens: A View into Next Token Information}
\label{logit-lens}
Logit lens technique, as first introduced by \cite{01-nostalgebraist2020logitlens}, provides a way for examining intermediate representations of model’s predictions at different stages of the model's output generation process, thereby providing insights into how the model’s internal representations evolve as model layers iteratively refine predictions. 

Specifically, given a pre-trained large language model (LLM), we use logit lens to decode a probability distribution over tokens from each intermediate layer. These token distributions represent model predictions after $l \in \{1, .., L\}$ layers of input processing. Given a sequence of tokens $t_1, .., t_n  \in V$, and $h_l^{(i)} \in R^d$  denoting the hidden state of token $t_i$ at layer $l$, the logits of the predictive distribution $p(t_{n+1}|t_1, .., t_n)$ are given by

\begin{equation}
[\text{logit}_1, ..., \text{logit}_{|V|}] = W_U \cdot \text{LayerNorm}_{L}(h_{L}^{(n)}),
\end{equation}

where $W_U \in R^{|V|\times d}$  denotes an embedding matrix, and $\text{LayerNorm}_{L}$ is the pre-embedding layer normalization. The logit lens applied the same unembedding operation to the earlier hidden states $h_{l}^{(i)}$:

\begin{equation}
[\text{logit}_1^{l}, ..., \text{logit}_{|V|}^{l}] = W_U \cdot \text{LayerNorm}_{L}(h_{l}^{(n)}),
\end{equation}

where an intermediate predictive distribution over tokens at layer $l$, $p_{l}(t_{n+1}|t_1, .., t_n)$, can be obtained.

\begin{figure*}[t]
    \centering
    \hfill
    \begin{subfigure}[h]{1\textwidth}
        \centering
        \includegraphics[width=1\textwidth]{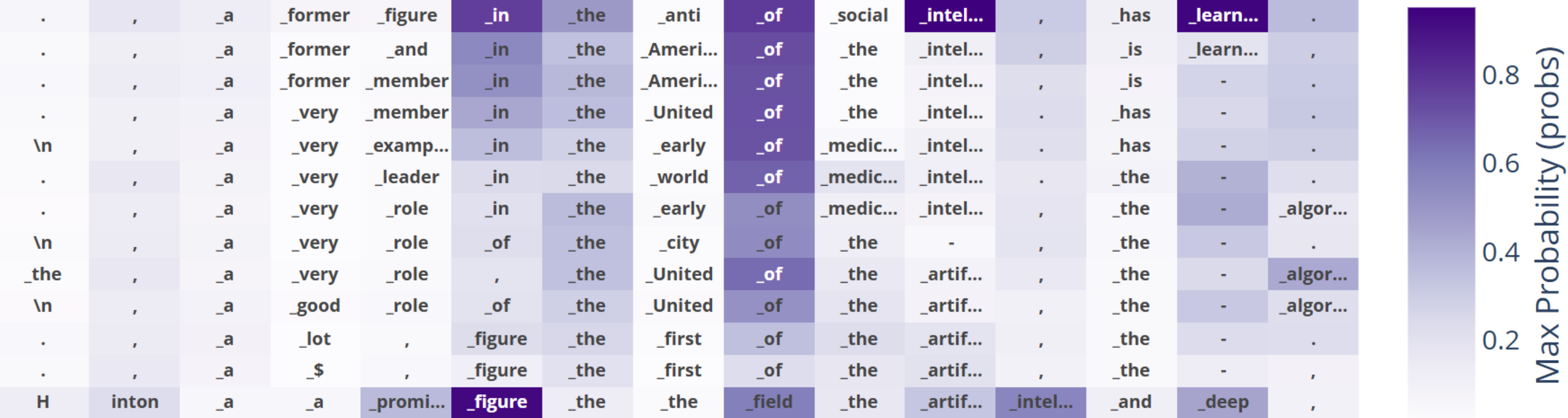}
        \caption{Maximum Probability}
    \end{subfigure}
    \hfill
    \hfill
    \vspace*{4mm}
    \begin{subfigure}[h]{1\textwidth}
        \centering
        \includegraphics[width=1\textwidth]{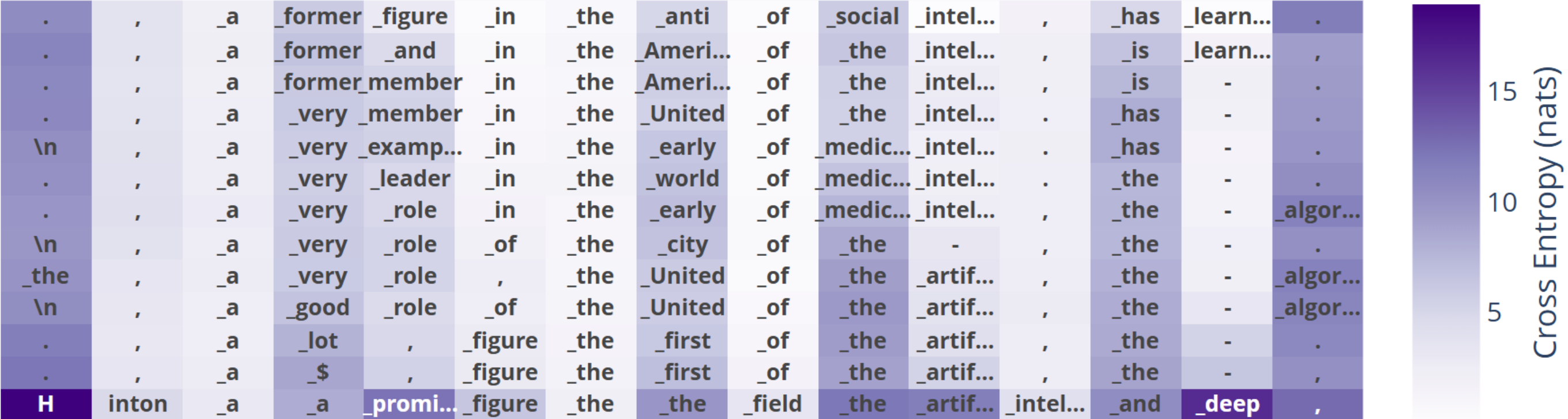}
        \caption{Cross-Entropy}
    \end{subfigure}
    \hfill
    \hfill
    \vspace*{4mm}
    \begin{subfigure}[h]{1\textwidth}
        \centering
        \includegraphics[width=1\textwidth]{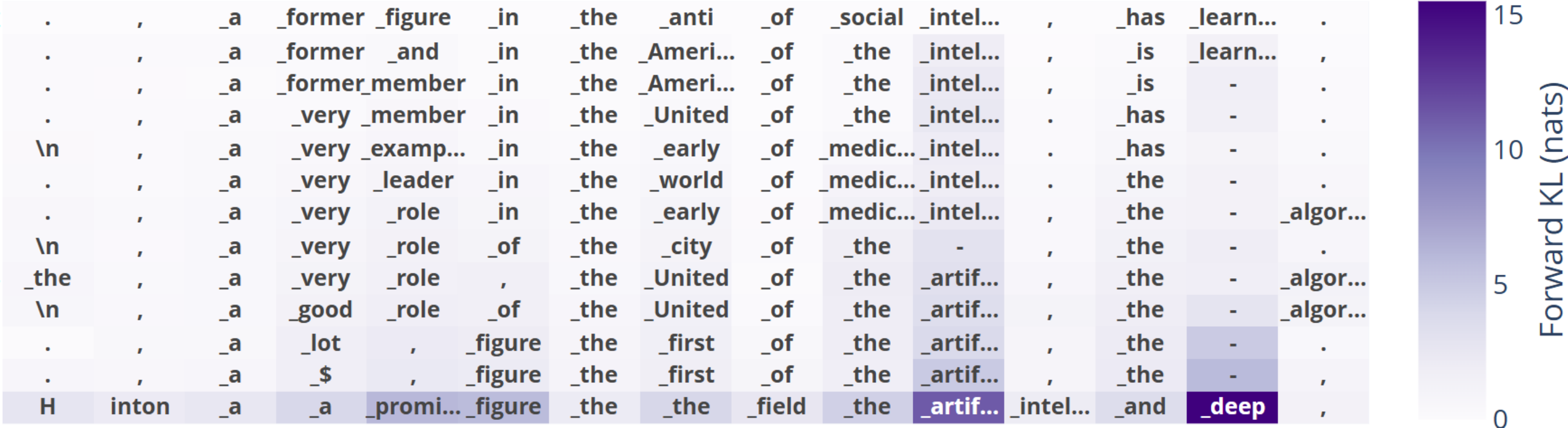}
        \caption{Forward Kullback-Leibler (KL) Divergence}
    \end{subfigure}
    \hfill
    \caption{The logit lens were applied to the hidden states of GPT-2 processing \textit{Hinton is a prominent figure in the field of artificial intelligence and deep learning.} The vertical axis represents layers ranging from 0 - 12 (bottom to top), and the horizon axis represents input tokens. Each cell illustrates the most likely next tokens (i.e., top-1 token prediction) in a sequence that the respective hidden state predicts. In case of maximum probability, the darker cells correspond to higher probabilities. And in case of cross-entropy and forward Kullback–Leibler (KL) divergence, the lighter cells correspond to higher probabilities.}
    \label{fig:logit-lens-main}
\end{figure*}

In this experiment, we utilized logit lens to hidden states of GPT-2 processing data of internet. For example, given an instance \textit{"Hinton is a prominent figure in the field of artificial intelligence and deep learning."}, shown in Figure \ref{fig:logit-lens-main}, we observed how distributions over next token predictions gradually converge to the final distribution. The observed, general trend is that initial predictions of tokens resemble plausible completions but seems far from the correct prediction. As layers progress, the predictions become more accurate with respect to the ground truth (next token). Roughly in the middle layers, a model forms a correct token of the next token, with later layers refining these predictions with higher predicted probabilities.

Take the \textit{6}th and \textit{9}th columns of Figure \ref{fig:logit-lens-main} as an example of the case where linking words (e.g., conjunctions and subordinating conjunctions), occur, in this case, \textit{"figure in"} and \textit{"field of"}. The model predicts the next word correctly in the early layer, without changing to other tokens in the later layers. And the model's confidence evolves over layers to the degree where the model is more confident about its final token prediction (nearly 100\% of predicted probabilities). Despite being trained on vast amounts of text without explicit syntactic rules, language models, GPT-2 in this case, show an ability to capture dependency relations, and structural aspects of language. Study reported by \citep{20-hewitt2019structural} shows that language models capture hierarchical syntactic structures, such as nested clauses, phrase boundaries, and subject-verb-object relationships. However, an intriguing question is how transformer-based models, trained solely on natural language data, manage to learn its hierarchical structure and generalize to sentences with unseen syntactic patterns?--despite lacking any explicitly encoded structural bias.

\citep{21-ahuja2024learning} explored the inductive biases in transformers that may drive this generalization. Through their extensive experiments on synthetic datasets with various training objectives, they found that, while objectives like sequence-to-sequence and prefix language modeling often fail to produce hierarchical generalization, models trained with the language modeling objective consistently are able to produce hierarchical generalization. The authors also conducted experiments on model pruning to examine how transformers encode hierarchical structure. They found (pruned) sub-networks with distinct generalization behaviors aligned with hierarchical structures and with linear order. Model pruning involves reducing the size of networks by removing unnecessary parameters while maintaining performance, Pruning models not only reduces model complexity and computational costs but also provides insights into model efficiency and interpretability. For example, pruning can reveal which parts of a network are most critical for certain tasks, shedding light on how models work internally.

Another interesting aspect is the model's demonstrated ability to grasp compositional understanding. For instance, in columns \textit{11}st and \textit{14}th, it recognizes how meaning arises from combining parts of a sentence, such as "artificial ... intelligence" and "deep ... learning," despite having multiple word choices that could follow "artificial" or "deep." Given the vast token space available at each layer, it is intriguing that the model not only selects the correct token but also reflects uncertainty in its decision with low probabilities. This somehow highlights its nuanced handling of language structure and meaning. However, this ability may not extend to more complex cases where the length of the input context and the position of relevant information vary. In such scenarios, as shown in Section \ref{input-context-logit-lens} for preliminary results, the model's performance may struggle to maintain the same level of compositional understanding and accuracy.

\begin{figure*}[t]
    \centering
    {
    \begin{subfigure}[b]{0.35\textwidth}
        \centering
        \includegraphics[width=\textwidth]{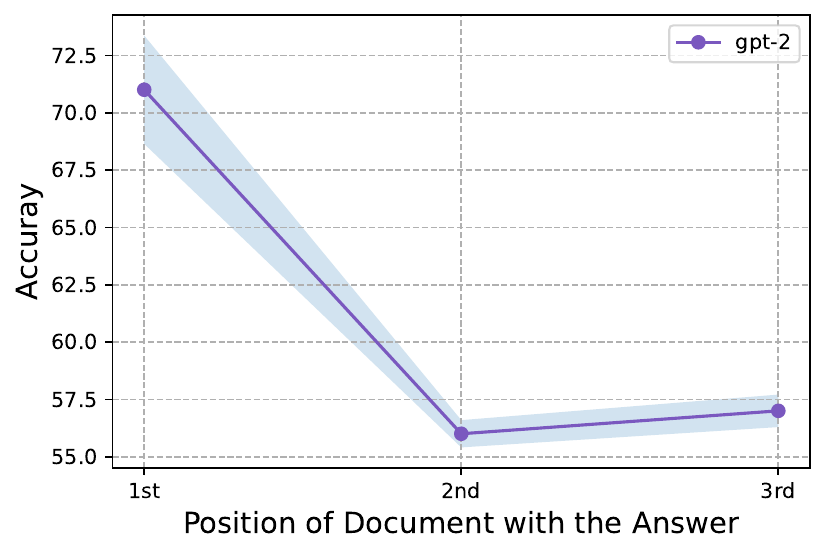}
    \end{subfigure}
    \begin{subfigure}[b]{0.64\textwidth}
        \centering
        \includegraphics[width=0.82\textwidth]{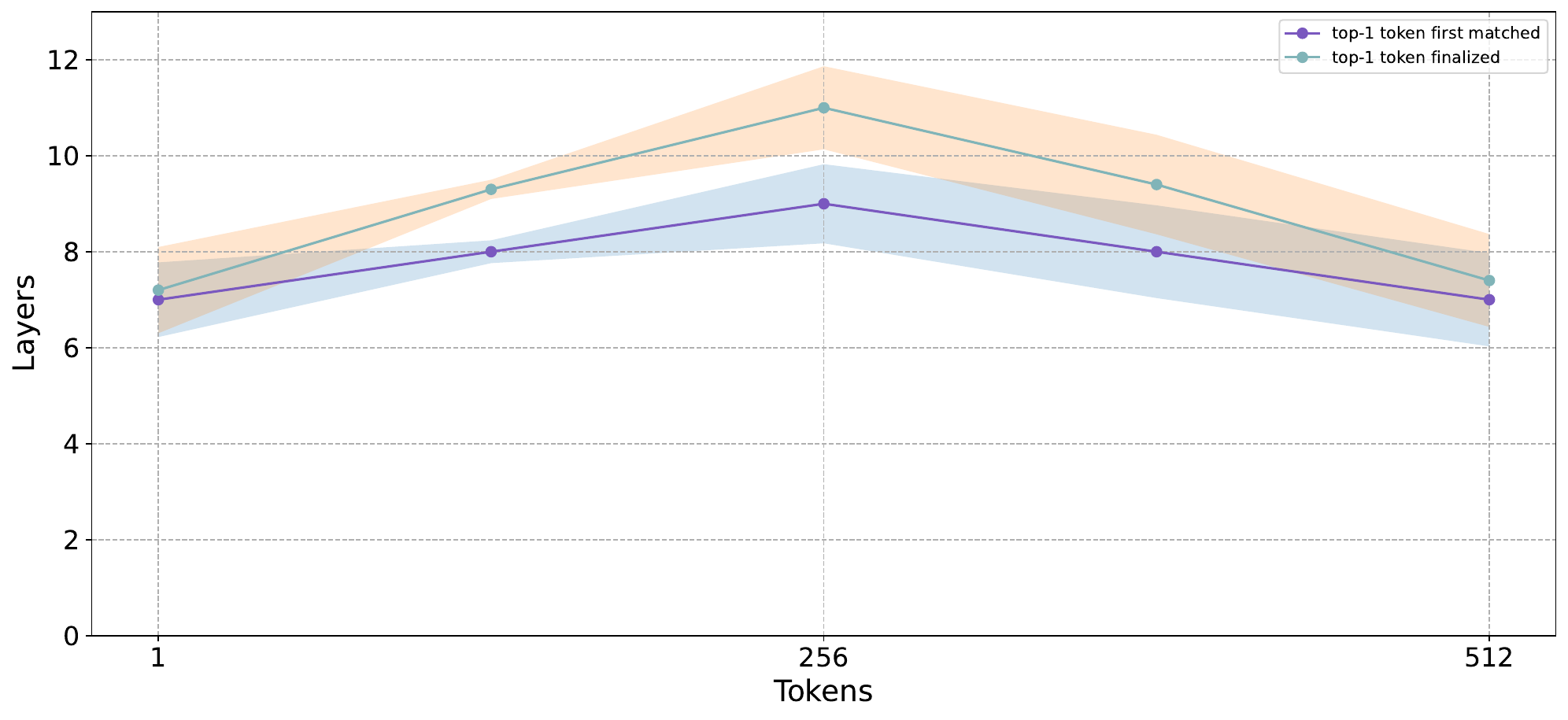}
    \end{subfigure}
    }
    
    {
     \begin{subfigure}[b]{0.35\textwidth}
        \centering
        \includegraphics[width=\textwidth]{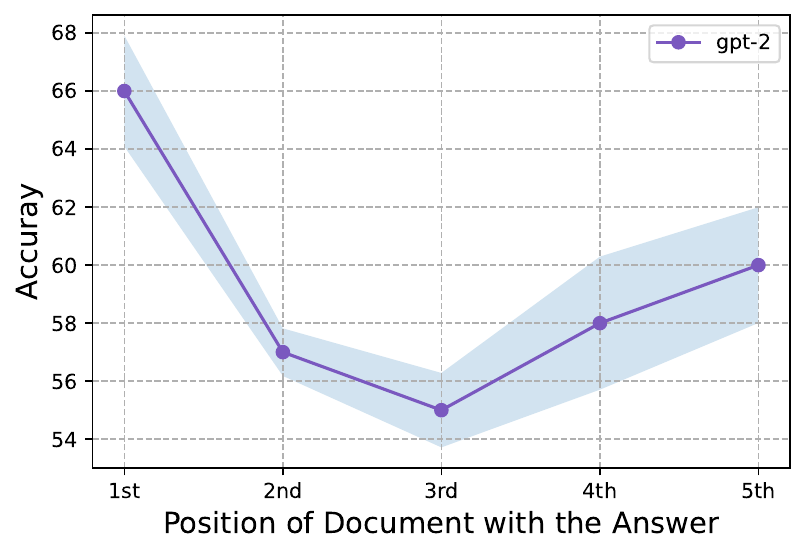}
    \end{subfigure}
    \begin{subfigure}[b]{0.64\textwidth}
        \centering
        \includegraphics[width=0.82\textwidth]{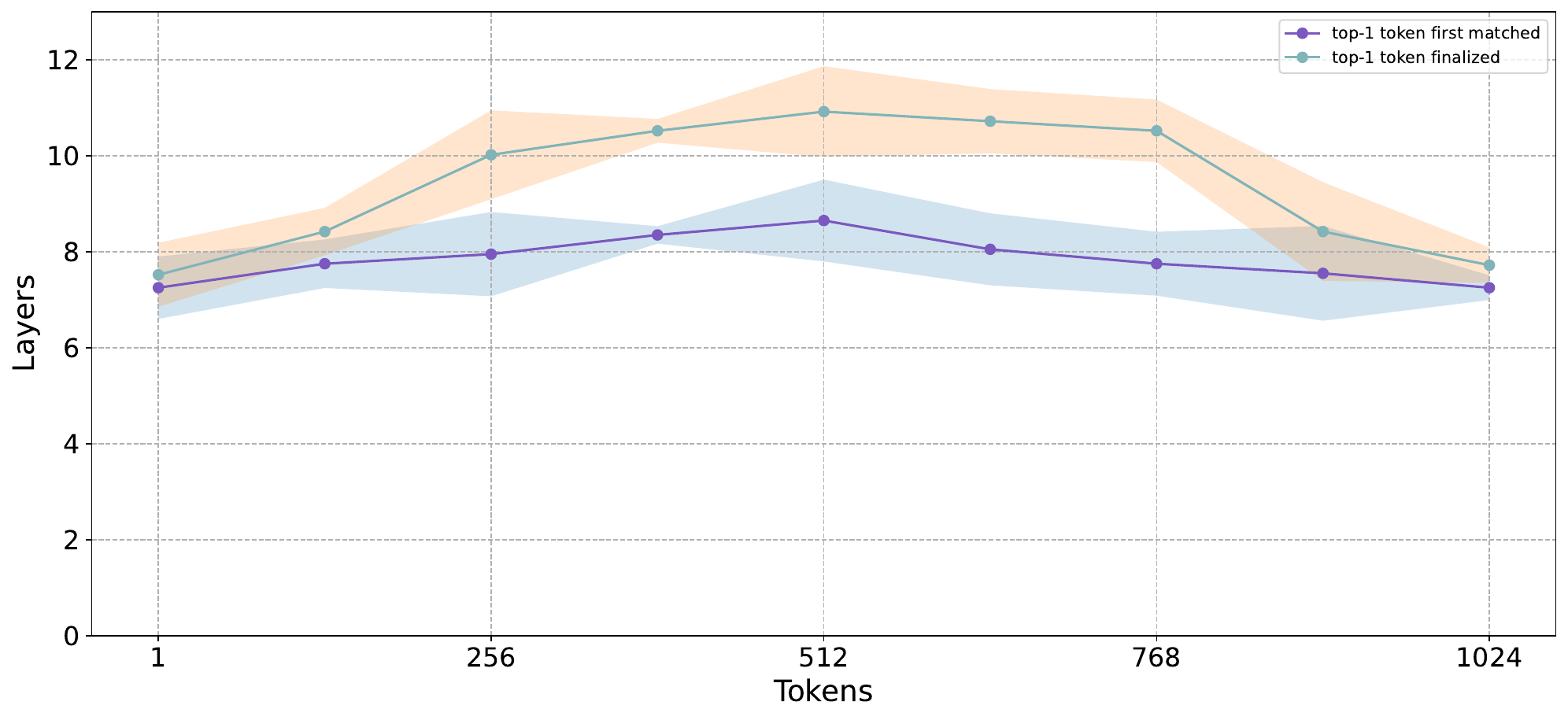}
    \end{subfigure}
    }
    
    \caption{Left: Shifting the location of relevant information within the model's input context reveals an U-shaped performance pattern. Model performs best when the relevant information is at the very beginning or end of the input, while its performance on average declines significantly when the critical information appears in the middle of the context. Right: Similarly, changing the position of relevant information produces a nearly inverted-U pattern, illustrating when the top-1 token first aligns with the correct answer and when it is finalized. The gap between these two layers on average is smaller when the relevant information is placed at the beginning or end of the input, indicating that the model needs more prediction refinements when processing long input context with the relevant information positioned in the middle. The experiments were repeated for 10 independent runs, and dots represent mean (arithmetic average) performance, with shaded bands represent 95\% confidence interval (CI).}

    \label{fig:context-length}
\end{figure*}

\section{Relation of Logit Lens to Input Context Lengths and the Position of Relevant Information within Input Context}
\label{input-context-logit-lens}

Input contexts for LLMs can contain thousands of tokens, especially when processing lengthy documents or integrating external information. For these tasks, LLMs must efficiently handle long sequences. Study done by \citep{09-ivgi2023efficient, 08-liu2024lost} explored how LLMs use input contexts to perform downstream tasks.

\citep{08-liu2024lost} designed an experiment to assess how LLMs access and use information from input contexts. They manipulated two factors: (1) the length of the input context and (2) the position of relevant information within it, to evaluate their impact on model performance. They hypothesized that if LLMs can reliably use information from long contexts, performance should remain stable regardless of the position of relevant information.

Multi-document question answering was selected for the experiments, where models must reason over multiple documents to extract relevant information and answer a question. This task reflects the retrieval-augmented generation process used in various applications. In more details about the set up, they controlled two key factors: (i) input context length by varying the number of documents (simulating different retrieval volumes), and (ii) the position of relevant information by altering the document order, placing the relevant one at the beginning, middle, or end of the context. Their findings showed that the position of relevant information in the input context significantly impacts model performance, revealing that current language models struggled to consistently access and use information in long contexts. Notably, they observed a U-shaped performance curve: models performed best when relevant information is at the beginning or the end of the input context, but their performance declines sharply when the information is located in the middle. In this experiment, we followed procedures and data sets used in \citep{08-liu2024lost}, to examine how token predictions are distributed as a function of input context length and position of relevant information.

In which layer(s) does the model’s correct prediction happen and finalize? As illustrated in Figure \ref{fig:context-length}, the top-1 token that matches the correct answer typically emerges in the middle or later layers. When handling long input contexts with relevant information placed in the middle, the correct top-1 tokens are identified similar to earlier, but it takes many additional layers before the model finalizes the prediction. The logit lens reveals how predictions are iteratively refined as they pass through each successive layer. Early layers may produce outputs that seem plausible but are far from accurate. The model begins with rough guesses, which are gradually improved as it incorporates more context and relevant information. This study indicates that when identifying bottlenecks or inefficiencies in the model, it's crucial to consider the position of relevant information. This positioning impacts to some extent the refinement process of token prediction, highlighting which layers on average are critical for determining the final output.

\begin{figure*}[t]
    \centering
    \hfill
    \begin{subfigure}[h]{1\textwidth}
        \centering
        \includegraphics[width=0.8\textwidth]{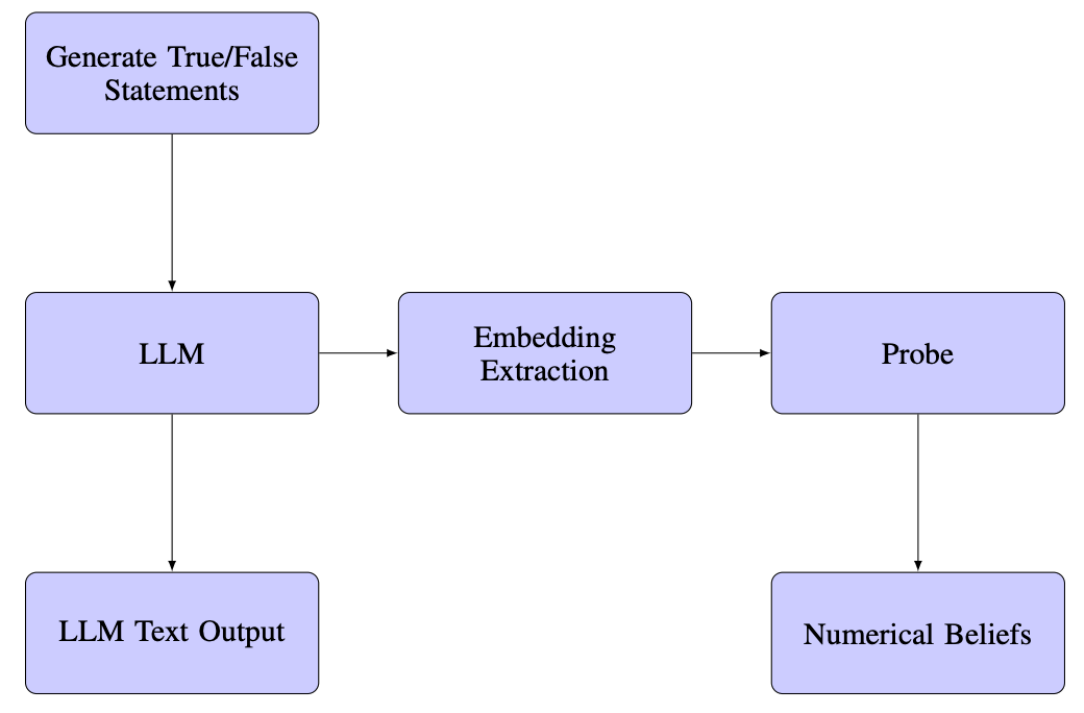}
    \end{subfigure}
    \hfill
    \caption{High-level overview of probing (Reproduced from Figure 2 in \citep{19-levinstein_still_2023}).}
    \label{fig:probing}
\end{figure*}

\section{Relation of Logit Lens to Probing }
\label{probing-logit-lens}

Probing refers to training a classifier to predict properties from internal representation to identify such properties, which is presumably encoded in learned representations \citep{07-alain2016understanding, 05-ettinger2016probing, 06-hupkes2018visualisation} (see Figure \ref{fig:probing}). High probing accuracy in a layer indicates that the correct answer can be extracted from its hidden states (see Section 4.1 in \citep{19-levinstein_still_2023}). However, this standard is often too easy to meet, particularly in straightforward classification tasks with high-dimensional hidden states \citep{02-hewitt2019designing}. In contrast, a high logit lens performance signifies that the layer effectively encodes correct answers along a direction in the residual stream, providing significantly more insight as shown in interventional experiments \citep{03-li2022emergent} that internal representations can be used to control the output of the network. Conversely, low logit lens performance does not necessarily mean that the correct answers cannot be decoded from that layer.

\section{Relevance to AI Safety}
\label{ai-safety}
Understanding inner working mechanisms of LLMs is essential for ensuring their safe development as we move toward more powerful models. Mechanistic interpretability approach has the potential to significantly advance LLM/AI safety research by providing a richer, stronger foundation for model evaluation \citep{casper_engineer_2023}. It can also offer early warnings of emergent capabilities, enabling us to understand better how internal structures and representations incrementally evolve as models learn, and to detect new skills or behaviors in models before they fully develop \citep{wei_emergent_2022, jacob_emergent_2023, nanda_progress_2023, barak_hidden_2022}. Furthermore, interpretability can strengthen theoretical risk models with concrete evidence, such as identifying inner misalignment (when a model’s behavior diverges from its intended goals). By exposing potential risks or problematic behaviors, interpretability could also prompt a shift within the AI community toward adopting more rigorous safety protocols \citep{hubinger_chris_2019}.

When it comes to specific AI risks \citep{hendrycks_overview_2023}, interpretability is a powerful tool for preventing malicious misuse by identifying and eliminating sensitive information embedded in models \citep{meng_locating_2022, nguyen_survey_2022}. It can also alleviate competitive pressures by providing clear evidence of potential threats, fostering a culture of safety within organizations, and reinforcing AI alignment—ensuring that AI systems stay aligned with intended goals—through enhanced monitoring and evaluation \citep{hendrycks_xrisk_2022}. In addition, interpretability offers essential safety checks throughout the AI development process. For example, prior to training, it informs deliberate design choices to enhance safety \citep{hubinger_chris_2019}; during training, it detects early signs of misalignment, enabling proactive shifts toward alignment \citep{hubinger_transparency_2022, sharkey_circumventing_2022}; and after training, it ensures rigorous evaluation of artificial cognition, verifying the model’s honesty \citep{burns_discovering_2023, zou_representation_2023} and screening for deceptive behaviors \citep{park_ai_2023}. This comprehensive approach helps safeguard AI systems at every stage, driving safer and more trustworthy advancements in AI.

The emergence of internal world models in LLMs holds transformative potential for AI alignment research. If we can identify internal representations of human values and align the AI system’s objectives accordingly, achieving alignment may become a straightforward task \citep{wentworth_how_2022}. This is especially promising if the AI's internal world model remains distinct from any notions of goals or agency \citep{ruthenis_internal_2022}. In such cases, simply interpreting the world model could be enough to ensure alignment \citep{ruthenis_worldmodel_2023}, providing an effective path to safer AI systems.

Mechanistic interpretability plays a vital role in advancing various AI alignment initiatives, including the understanding and control of existing models, facilitating AI systems in solving alignment challenges, and developing robust alignment theories \citep{technicalities_shallow_2023, hubinger_overview_2020}. By enhancing strategies to detect deceptive alignment—a scenario where a model appears aligned while pursuing misaligned goals without raising suspicion \citep{park_ai_2023}—and eliciting latent knowledge from models \citep{christiano_eliciting_2021}, we can significantly improve scalable oversight, such as through iterative distillation and amplification \citep{chan_what_2023}. Moreover, comprehensive interpretability can serve as an alignment strategy by helping us identify internal representations of human values and guiding the model to pursue those values through re-targeting an internal search process \citep{wentworth_how_2022}. Ultimately, the connection between understanding and control is crucial; a deeper understanding enables more reliable control over AI systems.


\section{Discussion and Limitations}
The effectiveness of the logit lens can vary significantly depending on the task and model used. For instance, tasks that require deep, abstract reasoning, layers far from the output may offer little relevant information when analyzed solely through token prediction.

Explaining how each layer of a model contributes to the final prediction by reducing complex representations to token predictions introduces biases. Many layers may serve other functions (e.g., refining internal features or re-weighting contextual information) which are not directly captured by logits. For instance, middle layers may not just be involved in predicting the next token, so using the logit lens alone may overlook their other roles in the model.

The logit lens relies on predicting which tokens have high probabilities at different layers. However, token distributions can be highly skewed, with some tokens dominating; even though they may not reflect the true nature of the underlying hidden state. This can obscure the true purpose or significance of intermediate representations.

Useful perspective on how language models construct their predictions can be obtained using logit lens. Its capabilities, however, are constrained when it comes to unveiling the model with more complex internal operations. Logit lens do not fully reveal the layered, nuanced, and abstract processes that underlie a language model's true capabilities. Thus it is essential to utilize logit lens in combination with other techniques that focus on different aspects of model behavior (e.g., neuron activity and attention patterns), to uncover the internal workings, and to obtain a deeper and more comprehensive understanding of how language models function.

\section{Conclusion}
Investigating the intermediate representations by examining how a model iteratively refines token predictions, offers insights into how a model's internal representations evolve. In this study, we utilized the logit lens to meticulously examine next token prediction refinements across different conditions, to some degree, to uncover critical nuances in how predictions are progressively refined. A deeper grasp of the prediction refinement process will enhance the development in AI safety research and development. That is, by establishing connections between the previous findings and existing hypotheses, we aim to advance the understanding of LLM behavior. These insights are vital for the development of AI technologies that are not only highly effective but also secure and aligned with human values.

\begin{ack}
The author would like to thank the support and opportunity given by Mukaya (Tai) Panich and Thanapong Boontaeng to research this topic. The author would also like to thank the members of the SCB 10X team for their useful comments during the contract term from May to October 2024.
\end{ack}

\newpage
\bibliographystyle{unsrtnat}
\bibliography{neurips_2024}

\begin{thebibliography}{50}
\providecommand{\natexlab}[1]{#1}
\providecommand{\url}[1]{\texttt{#1}}
\expandafter\ifx\csname urlstyle\endcsname\relax
  \providecommand{\doi}[1]{doi: #1}\else
  \providecommand{\doi}{doi: \begingroup \urlstyle{rm}\Url}\fi

\bibitem[Hendrycks and Mazeika(2022)]{hendrycks_xrisk_2022}
Dan Hendrycks and Mantas Mazeika.
\newblock X-risk analysis for ai research.
\newblock \emph{CoRR}, June 2022.
\newblock URL \url{https://arxiv.org/abs/2206.05862v7}.

\bibitem[Hendrycks et~al.(2023)Hendrycks, Mazeika, and Woodside]{hendrycks_overview_2023}
Dan Hendrycks, Mantas Mazeika, and Thomas Woodside.
\newblock An overview of catastrophic ai risks.
\newblock \emph{CoRR}, October 2023.
\newblock URL \url{http://arxiv.org/abs/2306.12001}.

\bibitem[Ji et~al.(2024)Ji, Qiu, Chen, Zhang, Lou, Wang, Duan, He, Zhou, Zhang, Zeng, Ng, Dai, Pan, O'Gara, Lei, Xu, Tse, Fu, McAleer, Yang, Wang, Zhu, Guo, and Gao]{ji_ai_2024}
Jiaming Ji, Tianyi Qiu, Boyuan Chen, Borong Zhang, Hantao Lou, Kaile Wang, Yawen Duan, Zhonghao He, Jiayi Zhou, Zhaowei Zhang, Fanzhi Zeng, Kwan~Yee Ng, Juntao Dai, Xuehai Pan, Aidan O'Gara, Yingshan Lei, Hua Xu, Brian Tse, Jie Fu, Stephen McAleer, Yaodong Yang, Yizhou Wang, Song-Chun Zhu, Yike Guo, and Wen Gao.
\newblock Ai alignment: A comprehensive survey.
\newblock \emph{CoRR}, January 2024.
\newblock \doi{10.48550/arXiv.2310.19852}.
\newblock URL \url{http://arxiv.org/abs/2310.19852}.

\bibitem[Olah(2022)]{olah_mechanistic_2022}
Christopher Olah.
\newblock Mechanistic interpretability, variables, and the importance of interpretable bases.
\newblock \emph{Transformer Circuits Thread}, 2022.
\newblock URL \url{https://transformer-circuits.pub/2022/mech-interp-essay/index.html}.

\bibitem[Nanda(2022)]{nanda_comprehensive_2022}
Neel Nanda.
\newblock A comprehensive mechanistic interpretability explainer \& glossary.
\newblock \emph{Neel Nanda's Blog}, December 2022.
\newblock URL \url{https://www.neelnanda.io/mechanistic-interpretability/glossary}.

\bibitem[Olah et~al.(2020)Olah, Cammarata, Schubert, Goh, Petrov, and Carter]{olah_zoom_2020}
Chris Olah, Nick Cammarata, Ludwig Schubert, Gabriel Goh, Michael Petrov, and Shan Carter.
\newblock Zoom in: An introduction to circuits.
\newblock \emph{Distill}, March 2020.
\newblock URL \url{https://distill.pub/2020/circuits/zoom-in}.

\bibitem[Sharkey et~al.(2022)Sharkey, Black, and {beren}]{sharkey_current_2022}
Lee Sharkey, Sid Black, and {beren}.
\newblock Current themes in mechanistic interpretability research.
\newblock \emph{AI Alignment Forum}, November 2022.
\newblock URL \url{https://www.alignmentforum.org/posts/Jgs7LQwmvErxR9BCC/current-themes-in-mechanistic-interpretability-research}.

\bibitem[Olah et~al.(2018)Olah, Satyanarayan, Johnson, Carter, Schubert, Ye, and Mordvintsev]{olah_building_2018}
Chris Olah, Arvind Satyanarayan, Ian Johnson, Shan Carter, Ludwig Schubert, Katherine Ye, and Alexander Mordvintsev.
\newblock The building blocks of interpretability.
\newblock \emph{Distill}, March 2018.
\newblock URL \url{https://distill.pub/2018/building-blocks}.

\bibitem[Nanda(2023)]{nanda_mechanistic_2023}
Neel Nanda.
\newblock Mechanistic interpretability quickstart guide.
\newblock \emph{Neel Nanda's Blog}, January 2023.
\newblock URL \url{https://www.neelnanda.io/mechanistic-interpretability/quickstart}.

\bibitem[Nanda(2024)]{nanda_extremely_2024}
Neel Nanda.
\newblock An extremely opinionated annotated list of my favourite mechanistic interpretability papers v2.
\newblock \emph{AI Alignment Forum}, July 2024.
\newblock URL \url{https://www.alignmentforum.org/posts/NfFST5Mio7BCAQHPA/an-extremely-opinionated-annotated-list-of-my-favourite}.

\bibitem[Das et~al.(2020)Das, Agarwal, Venugopal, Sheldon, and Shiva]{10-das2020taxonomy}
Saikat Das, Namita Agarwal, Deepak Venugopal, Frederick~T Sheldon, and Sajjan Shiva.
\newblock Taxonomy and survey of interpretable machine learning method.
\newblock In \emph{2020 IEEE Symposium Series on Computational Intelligence (SSCI)}, pages 670--677. IEEE, 2020.

\bibitem[Speith(2022)]{11-speith2022review}
Timo Speith.
\newblock A review of taxonomies of explainable artificial intelligence (xai) methods.
\newblock In \emph{Proceedings of the 2022 ACM conference on fairness, accountability, and transparency}, pages 2239--2250, 2022.

\bibitem[Bereska and Gavves(2024)]{12-bereska2024mechanistic}
Leonard Bereska and Efstratios Gavves.
\newblock Mechanistic interpretability for ai safety--a review.
\newblock \emph{arXiv preprint arXiv:2404.14082}, 2024.

\bibitem[Zytek et~al.(2022)Zytek, Arnaldo, Liu, Berti-Equille, and Veeramachaneni]{13-zytek2022need}
Alexandra Zytek, Ignacio Arnaldo, Dongyu Liu, Laure Berti-Equille, and Kalyan Veeramachaneni.
\newblock The need for interpretable features: Motivation and taxonomy.
\newblock \emph{ACM SIGKDD Explorations Newsletter}, 24\penalty0 (1):\penalty0 1--13, 2022.

\bibitem[Zhang et~al.(2021)Zhang, Ti{\v{n}}o, Leonardis, and Tang]{14-zhang2021survey}
Yu~Zhang, Peter Ti{\v{n}}o, Ale{\v{s}} Leonardis, and Ke~Tang.
\newblock A survey on neural network interpretability.
\newblock \emph{IEEE Transactions on Emerging Topics in Computational Intelligence}, 5\penalty0 (5):\penalty0 726--742, 2021.

\bibitem[Alain(2016)]{07-alain2016understanding}
Guillaume Alain.
\newblock Understanding intermediate layers using linear classifier probes.
\newblock \emph{arXiv preprint arXiv:1610.01644}, 2016.

\bibitem[Ettinger et~al.(2016)Ettinger, Elgohary, and Resnik]{05-ettinger2016probing}
Allyson Ettinger, Ahmed Elgohary, and Philip Resnik.
\newblock Probing for semantic evidence of composition by means of simple classification tasks.
\newblock In \emph{Proceedings of the 1st workshop on evaluating vector-space representations for nlp}, pages 134--139, 2016.

\bibitem[Hupkes et~al.(2018)Hupkes, Veldhoen, and Zuidema]{06-hupkes2018visualisation}
Dieuwke Hupkes, Sara Veldhoen, and Willem Zuidema.
\newblock Visualisation and'diagnostic classifiers' reveal how recurrent and recursive neural networks process hierarchical structure.
\newblock \emph{Journal of Artificial Intelligence Research}, 61:\penalty0 907--926, 2018.

\bibitem[Burns et~al.(2022)Burns, Ye, Klein, and Steinhardt]{18-burns2022discovering}
Collin Burns, Haotian Ye, Dan Klein, and Jacob Steinhardt.
\newblock Discovering latent knowledge in language models without supervision.
\newblock \emph{arXiv preprint arXiv:2212.03827}, 2022.

\bibitem[Cunningham et~al.(2023)Cunningham, Ewart, Riggs, Huben, and Sharkey]{17-cunningham2023sparse}
Hoagy Cunningham, Aidan Ewart, Logan Riggs, Robert Huben, and Lee Sharkey.
\newblock Sparse autoencoders find highly interpretable features in language models.
\newblock \emph{arXiv preprint arXiv:2309.08600}, 2023.

\bibitem[Wang et~al.(2022)Wang, Variengien, Conmy, Shlegeris, and Steinhardt]{15-wang2022interpretability}
Kevin Wang, Alexandre Variengien, Arthur Conmy, Buck Shlegeris, and Jacob Steinhardt.
\newblock Interpretability in the wild: a circuit for indirect object identification in gpt-2 small.
\newblock \emph{arXiv preprint arXiv:2211.00593}, 2022.

\bibitem[Goldowsky-Dill et~al.(2023)Goldowsky-Dill, MacLeod, Sato, and Arora]{16-goldowsky2023localizing}
Nicholas Goldowsky-Dill, Chris MacLeod, Lucas Sato, and Aryaman Arora.
\newblock Localizing model behavior with path patching.
\newblock \emph{arXiv preprint arXiv:2304.05969}, 2023.

\bibitem[nostalgebraist(2020)]{01-nostalgebraist2020logitlens}
nostalgebraist.
\newblock interpreting gpt: the logit lens.
\newblock \emph{LessWrong}, 2020.
\newblock URL \url{https://www.lesswrong.com/posts/AcKRB8wDpdaN6v6ru/interpreting-gpt-the-logit-lens}.

\bibitem[Hewitt and Manning(2019)]{20-hewitt2019structural}
John Hewitt and Christopher~D Manning.
\newblock A structural probe for finding syntax in word representations.
\newblock In \emph{Proceedings of the 2019 Conference of the North American Chapter of the Association for Computational Linguistics: Human Language Technologies, Volume 1 (Long and Short Papers)}, pages 4129--4138, 2019.

\bibitem[Ahuja et~al.(2024)Ahuja, Balachandran, Panwar, He, Smith, Goyal, and Tsvetkov]{21-ahuja2024learning}
Kabir Ahuja, Vidhisha Balachandran, Madhur Panwar, Tianxing He, Noah~A Smith, Navin Goyal, and Yulia Tsvetkov.
\newblock Learning syntax without planting trees: Understanding when and why transformers generalize hierarchically.
\newblock \emph{arXiv preprint arXiv:2404.16367}, 2024.

\bibitem[Ivgi et~al.(2023)Ivgi, Shaham, and Berant]{09-ivgi2023efficient}
Maor Ivgi, Uri Shaham, and Jonathan Berant.
\newblock Efficient long-text understanding with short-text models.
\newblock \emph{Transactions of the Association for Computational Linguistics}, 11:\penalty0 284--299, 2023.

\bibitem[Liu et~al.(2024)Liu, Lin, Hewitt, Paranjape, Bevilacqua, Petroni, and Liang]{08-liu2024lost}
Nelson~F Liu, Kevin Lin, John Hewitt, Ashwin Paranjape, Michele Bevilacqua, Fabio Petroni, and Percy Liang.
\newblock Lost in the middle: How language models use long contexts.
\newblock \emph{Transactions of the Association for Computational Linguistics}, 12:\penalty0 157--173, 2024.

\bibitem[Levinstein and Herrmann(2023)]{19-levinstein_still_2023}
B.~A. Levinstein and Daniel~A. Herrmann.
\newblock Still no lie detector for language models: Probing empirical and conceptual roadblocks.
\newblock \emph{CoRR}, June 2023.
\newblock \doi{10.48550/arXiv.2307.00175}.
\newblock URL \url{http://arxiv.org/abs/2307.00175}.

\bibitem[Hewitt and Liang(2019)]{02-hewitt2019designing}
John Hewitt and Percy Liang.
\newblock Designing and interpreting probes with control tasks.
\newblock \emph{arXiv preprint arXiv:1909.03368}, 2019.

\bibitem[Li et~al.(2022)Li, Hopkins, Bau, Vi{\'e}gas, Pfister, and Wattenberg]{03-li2022emergent}
Kenneth Li, Aspen~K Hopkins, David Bau, Fernanda Vi{\'e}gas, Hanspeter Pfister, and Martin Wattenberg.
\newblock Emergent world representations: Exploring a sequence model trained on a synthetic task.
\newblock \emph{arXiv preprint arXiv:2210.13382}, 2022.

\bibitem[Casper(2023)]{casper_engineer_2023}
Stephen Casper.
\newblock The engineer's interpretability sequence.
\newblock \emph{AI Alignment Forum}, February 2023.
\newblock URL \url{https://www.alignmentforum.org/s/a6ne2ve5uturEEQK7}.

\bibitem[Wei et~al.(2022)Wei, Tay, Bommasani, Raffel, Zoph, Borgeaud, Yogatama, Bosma, Zhou, Metzler, Chi, Hashimoto, Vinyals, Liang, Dean, and Fedus]{wei_emergent_2022}
Jason Wei, Yi~Tay, Rishi Bommasani, Colin Raffel, Barret Zoph, Sebastian Borgeaud, Dani Yogatama, Maarten Bosma, Denny Zhou, Donald Metzler, Ed~H. Chi, Tatsunori Hashimoto, Oriol Vinyals, Percy Liang, Jeff Dean, and William Fedus.
\newblock Emergent abilities of large language models.
\newblock \emph{TMLR}, October 2022.
\newblock \doi{10.48550/arXiv.2206.07682}.
\newblock URL \url{http://arxiv.org/abs/2206.07682}.

\bibitem[Steinhardt(2023)]{jacob_emergent_2023}
Jacob Steinhardt.
\newblock Emergent deception and emergent optimization.
\newblock \emph{Bounded Regret}, February 2023.
\newblock URL \url{https://bounded-regret.ghost.io/emergent-deception-optimization/}.

\bibitem[Nanda et~al.(2023)Nanda, Chan, Lieberum, Smith, and Steinhardt]{nanda_progress_2023}
Neel Nanda, Lawrence Chan, Tom Lieberum, Jess Smith, and Jacob Steinhardt.
\newblock Progress measures for grokking via mechanistic interpretability.
\newblock \emph{ICLR}, January 2023.
\newblock \doi{10.48550/arXiv.2301.05217}.
\newblock URL \url{http://arxiv.org/abs/2301.05217}.

\bibitem[Barak et~al.(2022)Barak, Edelman, Goel, Kakade, Malach, and Zhang]{barak_hidden_2022}
Boaz Barak, Benjamin~L. Edelman, Surbhi Goel, Sham Kakade, Eran Malach, and Cyril Zhang.
\newblock Hidden progress in deep learning: Sgd learns parities near the computational limit.
\newblock \emph{NeurIPS}, 2022.
\newblock \doi{10.48550/arXiv.2207.08799}.
\newblock URL \url{http://arxiv.org/abs/2207.08799}.

\bibitem[Hubinger(2019)]{hubinger_chris_2019}
Evan Hubinger.
\newblock Chris olah's views on agi safety.
\newblock \emph{AI Alignment Forum}, November 2019.
\newblock URL \url{https://www.alignmentforum.org/posts/X2i9dQQK3gETCyqh2/chris-olah-s-views-on-agi-safety}.

\bibitem[Meng et~al.(2022)Meng, Bau, Andonian, and Belinkov]{meng_locating_2022}
Kevin Meng, David Bau, Alex Andonian, and Yonatan Belinkov.
\newblock Locating and editing factual associations in gpt.
\newblock \emph{NeurIPS}, 2022.
\newblock \doi{10.48550/arXiv.2202.05262}.
\newblock URL \url{http://arxiv.org/abs/2202.05262}.

\bibitem[Nguyen et~al.(2022)Nguyen, Huynh, Nguyen, Liew, Yin, and Nguyen]{nguyen_survey_2022}
Thanh~Tam Nguyen, Thanh~Trung Huynh, Phi~Le Nguyen, Alan Wee-Chung Liew, Hongzhi Yin, and Quoc Viet~Hung Nguyen.
\newblock A survey of machine unlearning.
\newblock \emph{CoRR}, October 2022.
\newblock \doi{10.48550/arXiv.2209.02299}.
\newblock URL \url{http://arxiv.org/abs/2209.02299}.

\bibitem[Hubinger(2022)]{hubinger_transparency_2022}
Evan Hubinger.
\newblock A transparency and interpretability tech tree.
\newblock \emph{AI Alignment Forum}, June 2022.
\newblock URL \url{https://www.alignmentforum.org/posts/nbq2bWLcYmSGup9aF/a-transparency-and-interpretability-tech-tree}.

\bibitem[Sharkey(2022)]{sharkey_circumventing_2022}
Lee Sharkey.
\newblock Circumventing interpretability: How to defeat mind-readers.
\newblock \emph{CoRR}, December 2022.
\newblock \doi{10.48550/ARXIV.2212.11415}.
\newblock URL \url{https://arxiv.org/abs/2212.11415}.

\bibitem[Burns et~al.(2023)Burns, Ye, Klein, and Steinhardt]{burns_discovering_2023}
Collin Burns, Haotian Ye, Dan Klein, and Jacob Steinhardt.
\newblock Discovering latent knowledge in language models without supervision.
\newblock \emph{ICLR}, 2023.
\newblock URL \url{http://arxiv.org/abs/2212.03827}.

\bibitem[Zou et~al.(2023)Zou, Phan, Chen, Campbell, Guo, Ren, Pan, Yin, Mazeika, Dombrowski, Goel, Li, Byun, Wang, Mallen, Basart, Koyejo, Song, Fredrikson, Kolter, and Hendrycks]{zou_representation_2023}
Andy Zou, Long Phan, Sarah Chen, James Campbell, Phillip Guo, Richard Ren, Alexander Pan, Xuwang Yin, Mantas Mazeika, Ann-Kathrin Dombrowski, Shashwat Goel, Nathaniel Li, Michael~J. Byun, Zifan Wang, Alex Mallen, Steven Basart, Sanmi Koyejo, Dawn Song, Matt Fredrikson, J.~Zico Kolter, and Dan Hendrycks.
\newblock Representation engineering: A top-down approach to ai transparency.
\newblock \emph{CoRR}, October 2023.
\newblock \doi{10.48550/arXiv.2310.01405}.
\newblock URL \url{http://arxiv.org/abs/2310.01405}.

\bibitem[Park et~al.(2023)Park, Goldstein, O'Gara, Chen, and Hendrycks]{park_ai_2023}
Peter~S. Park, Simon Goldstein, Aidan O'Gara, Michael Chen, and Dan Hendrycks.
\newblock Ai deception: A survey of examples, risks, and potential solutions.
\newblock \emph{CoRR}, August 2023.
\newblock \doi{10.48550/arXiv.2308.14752}.
\newblock URL \url{http://arxiv.org/abs/2308.14752}.

\bibitem[Wentworth(2022)]{wentworth_how_2022}
John Wentworth.
\newblock How to go from interpretability to alignment: Just retarget the search.
\newblock \emph{AI Alignment Forum}, August 2022.
\newblock URL \url{https://www.alignmentforum.org/posts/w4aeAFzSAguvqA5qu/how-to-go-from-interpretability-to-alignment-just-retarget}.

\bibitem[Ruthenis(2022)]{ruthenis_internal_2022}
Thane Ruthenis.
\newblock Internal interfaces are a high-priority interpretability target.
\newblock \emph{AI Alignment Forum}, December 2022.
\newblock URL \url{https://www.lesswrong.com/posts/nwLQt4e7bstCyPEXs/internal-interfaces-are-a-high-priority-interpretability}.

\bibitem[Ruthenis(2023)]{ruthenis_worldmodel_2023}
Thane Ruthenis.
\newblock World-model interpretability is all we need.
\newblock \emph{AI Alignment Forum}, January 2023.
\newblock URL \url{https://www.alignmentforum.org/posts/HaHcsrDSZ3ZC2b4fK/world-model-interpretability-is-all-we-need}.

\bibitem[{technicalities} and Stag(2023)]{technicalities_shallow_2023}
{technicalities} and Stag.
\newblock Shallow review of live agendas in alignment \& safety.
\newblock \emph{LessWrong}, 2023.
\newblock URL \url{https://www.lesswrong.com/posts/zaaGsFBeDTpCsYHef/shallow-review-of-live-agendas-in-alignment-and-safety}.

\bibitem[Hubinger(2020)]{hubinger_overview_2020}
Evan Hubinger.
\newblock An overview of 11 proposals for building safe advanced ai.
\newblock \emph{CoRR}, December 2020.
\newblock \doi{10.48550/arXiv.2012.07532}.
\newblock URL \url{http://arxiv.org/abs/2012.07532}.

\bibitem[Christiano et~al.(2021)Christiano, Cotra, and Xu]{christiano_eliciting_2021}
Paul Christiano, Ajeya Cotra, and Mark Xu.
\newblock Eliciting latent knowledge, January 2021.
\newblock URL \url{https://docs.google.com/document/d/1WwsnJQstPq91_Yh-Ch2XRL8H_EpsnjrC1dwZXR37PC8/edit?usp=sharing&usp=embed_facebook}.

\bibitem[Chan(2023)]{chan_what_2023}
Lawrence Chan.
\newblock What i would do if i wasn't at arc evals.
\newblock \emph{AI Alignment Forum}, May 2023.
\newblock URL \url{https://www.lesswrong.com/posts/6FkWnktH3mjMAxdRT/what-i-would-do-if-i-wasn-t-at-arc-evals}.

\end{thebibliography}


\end{document}